\documentclass[10pt,twocolumn,letterpaper]{article}

\usepackage{cvpr}              

\usepackage{amsmath}
\usepackage{amssymb}
\usepackage{graphicx}
\usepackage{subcaption}
\usepackage{amsthm}
\newtheorem{theorem}{Theorem}
\usepackage{tikz}
\usetikzlibrary{positioning}
\usetikzlibrary{arrows.meta}
\usepackage{adjustbox}
\usepackage{float}
\usepackage{placeins}

\usepackage{fancyhdr}
\fancypagestyle{myfooter}{
  \fancyhf{} 
  \cfoot{\footnotesize CVPR 2025 Tutorial - Identifying Structure in Data: All you need to know about Dimensionality Reduction, Clustering, and More	 }
  \renewcommand{\headrulewidth}{0pt} 
  \renewcommand{\footrulewidth}{0pt}
}

\definecolor{cvprblue}{rgb}{0.21,0.49,0.74}
\usepackage[pagebackref,breaklinks,colorlinks,allcolors=cvprblue]{hyperref}

\usepackage{float}
\usepackage{stfloats}

\def\paperID{1234}
\def\confName{CVPR}
\def\confYear{2025}

\title{NOMAD Projection}

\author{Brandon Duderstadt \\
Nomic AI\\
{\tt\small brandon@nomic.ai}
\and
Zach Nussbaum\\
Nomic AI\\
{\tt\small zach@nomic.ai}
\and
Laurens van der Maaten\\
Nomic AI\\
{\tt\small lvdmaaten@gmail.com}
}

\begin{document}
\maketitle

\begin{abstract}
The rapid adoption of generative AI has driven an explosion in the size of datasets consumed and produced by AI models.
Traditional methods for unstructured data visualization, such as t-SNE and UMAP, have not kept up with the pace of dataset scaling.
This presents a significant challenge for AI explainability, which relies on methods such as t-SNE and UMAP for exploratory data analysis.
In this paper, we introduce Negative Or Mean Affinity Discrimination (NOMAD) Projection, the first method for unstructured data visualization via nonlinear dimensionality reduction that can run on multiple GPUs at train time.
We provide theory that situates NOMAD Projection as an approximate upper bound on the InfoNC-t-SNE loss, and empirical results that demonstrate NOMAD Projection's superior performance and speed profile compared to existing state-of-the-art methods.
We demonstrate the scalability of NOMAD Projection by computing the first complete data map of Multilingual Wikipedia.
\end{abstract}

\thispagestyle{myfooter}

\section{Introduction}
The discovery of neural scaling laws has resulted in an explosion in the size of datasets consumed and produced by AI models \citep{kaplan2020scalinglawsneurallanguage} \cite{hoffmann2022trainingcomputeoptimallargelanguage}.
Traditional algorithms for unstructured data visualization, such as t-SNE \citep{maaten2008visualizing} and UMAP \citep{mcinnes2020umapuniformmanifoldapproximation}, have not kept up with the pace of dataset scaling.
The presents a significant challenge for data-centric AI explainability, since it relies upon methods like t-SNE and UMAP for exploratory data analysis.

Traditional unstructured data visualization methods scale very poorly in both wall-clock time and memory usage.
Both these scaling issues can be resolved by moving towards a massively distributed paradigm for computing the visualizations, akin to how deep learning has moved to a distributed training paradigm.
Distributing traditional unstructured data visualization algorithms is a non-trivial problem since their loss functions do not factorize across data points, making multi-device orchestration challenging.

In this paper, we introduce Negative Or Mean Affinity Discrimination (NOMAD) Projection, an unstructured data visualization method that enables distributed computation by approximating an upper bound on the InfoNC-t-SNE loss.
We demonstrate that NOMAD Projection effectively scales to multiple GPUs, which significantly reduces wall-clock time and enables us to quickly compute visualizations of datasets that do not fit into the memory of a single GPU.

\section{Background}

Nonlinear dimensionality reduction for use in data visualization has been a focus of the machine learning community for over a decade.
There has been a particular focus on the data mapping problem, which uses nonlinear dimensionality reduction to project vector data to the 2D plane so it can be rendered on a screen for subsequent analysis.

Early algorithms for solving the data mapping problem include classical multidimensional scaling \citep{Torgerson1952}, Sammon mapping \cite{Sammon1969}, Locally Linear Embeddings \citep{roweis2000nonlinear},
Isomap \citep{tenenbaum2000global}, and Stochastic Neighbor Embedding (SNE; \citet{hinton2002stochastic}.
The first method to gain widespread adoption was t-SNE \cite{maaten2008visualizing}, which solved the crowding problem present in SNE.

It was discovered that data mapping algorithms scaled poorly in both wall-time and memory consumption with the size of the input data.
Several methods, including Barnes-Hut t-SNE \cite{vandermaaten2013barneshutsne}, LargeVis \citep{Tang_2016}, UMAP \citep{mcinnes2020umapuniformmanifoldapproximation}, and NCVis \citep{artemenkov2020ncvisnoisecontrastiveapproach}  leverage approximations in an attempt to improve their scalability.

Work by \citet{damrich2023tsneumapcontrastivelearning} has unified the design choices of t-SNE, LargeVis, UMAP, and NCVis into a cohesive theoretical framework using the language of contrastive learning.
Under their contrastive framework, solving the data mapping problem is interpreted as iteratively optimizing a spring system, where positive contrastive pairs are connected by attractive spring forces and negative contrastive pairs are connected by repulsive spring forces. 

Naive solutions to the data mapping problem compute positive and negative forces between all pairs of points, leading to quadratic complexity and poor scalability.
LargeVis, UMAP, and NCVis all use similar approximation methods to sidestep the naive quadratic complexity and improve their scalability.
In particular, they employ a kNN graph to compute a linear number of positive spring forces for each point, and use negative sampling or noise contrastive estimation (NCE) \citep{ma2018noisecontrastiveestimationnegative} to compute a linear number of negative forces for each point.

\citet{damrich2023tsneumapcontrastivelearning} use the connection between data mapping and contrastive learning to derive InfoNC-t-SNE, a NCE approximation to t-SNE based on the InfoNCE loss \citep{DBLP:journals/corr/abs-1807-03748}.
In InfoNC-t-SNE, edges in a kNN graph constructed from a vector dataset correspond to positive contrastive pairs, and edges from a complete graph over the dataset correspond to negative contrastive pairs. During gradient based optimization of the InfoNC-t-SNE loss, positive contrastive pairs induce attractive spring forces between points, and negative contrastive pairs induce repulsive spring forces between points.

Alongside algorithmic advances, implementations such as t-SNE-cuda \citep{chan2018tsnecudagpuacceleratedtsneapplications} and RapidsUMAP \citep{nolet2021bringingumapcloserspeed} have improved data mapping scalability by taking advantage of powerful GPU hardware.
However, none of these implementations can effectively utilize multiple GPUs at train time.
This severely limits their scalability, as low GPU vRAM caps bottleneck the size of data they can be run on.

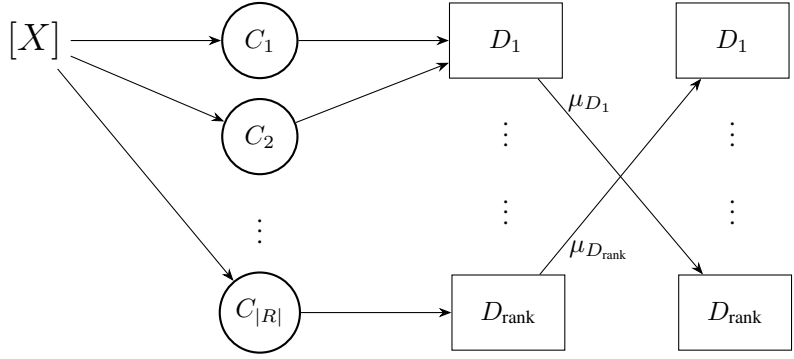
\begin{figure*}
    \centering
    \begin{tikzpicture}[
        block/.style={rectangle, draw, minimum width=1.5cm, minimum height=1cm},
        arrow/.style={-{Stealth[]}},
        node distance=2cm and 2cm,
        large_matrix/.style={minimum width=2.5cm, minimum height=1.5cm, draw, thick},
        circ/.style={circle, draw, minimum size=1.0cm, text centered, thick}
    ]

    \node (x) at (0,0) {\textbf{\Large $[X]$}};

    \node (c1) [right=2cm of x, circ] {$C_1$};
    \node (c2) [below=0.25cm of c1, circ] {$C_2$};
    \node (dots) [below=0.25cm of c2] {$\vdots$};
    \node (cp) [below=0.25cm of dots, circ] {$C_{|R|}$};

    \node (d1) [right=of c1, block] {$D_1$};
    \node (dots2) [below=0.25cm of d1] {$\vdots$};
    \node (dots3) [below=0.25cm of dots2] {$\vdots$};
    \node (dp) [right=of cp, block] {$D_\text{rank}$};

    \node (d1_target) [right=1.5cm of d1, block] {$D_1$};
    \node (dp_target) [right=1.5cm of dp, block] {$D_\text{rank}$};
    \node (dots4) [below=0.25cm of d1_target] {$\vdots$};
    \node (dots5) [below=0.25cm of dots4] {$\vdots$};

    \draw [arrow] (x) -- (c1);
    \draw [arrow] (c2) -- (d1);
    \draw [arrow] (x) -- (c2);
    \draw [arrow] (x) -- (cp);

    \draw [arrow] (c1) -- (d1);
    \draw [arrow] (cp) -- (dp);

    \draw [arrow] (d1) -- (dp_target) node[very near start, right] {$\mu_{D_1}$};
    \draw [arrow] (dp) -- (d1_target) node[very near start, right] {$\mu_{D_\text{rank}}$};

    \end{tikzpicture}

    \caption{NOMAD Projection's Distributed Training Strategy - First, input data is partitioned into clusters $C_1, C_2, ..., C_{|R|}$ during the creation of the ANN index. Clusters are then sharded across devices $D_1, ..., D_{\text{rank}}$. Since each cluster is a component of the ANN graph, no inter-device communication is required during positive spring force calculation. After every epoch, only the matrices of cluster means $\mu_{D_1}, ..., \mu_{D_\text{rank}}$ are all-gathered, minimizing the inter-device communication required for negative spring force calculation.}
    \label{fig:distribution}
\end{figure*}

\section{NOMAD Projection}

In this section, we develop Negative Or Mean Affinity Discrimination (NOMAD)
Projection.
NOMAD Projection approximates an upper bound on the InfoNC-t-SNE loss, enabling it to easily scale to multiple GPUs at train time. 

\subsection{InfoNC-t-SNE}
InfoNC-t-SNE learns low dimensional positions for high dimensional data by framing the problem as classification. In particular, InfoNC-t-SNE learns a classifier that can discriminate between true edges in a kNN graph constructed over vector data and randomly sampled noise edges from a complete variant of the kNN graph.
Importantly, the InfoNC-t-SNE classifier is parameterized by the low dimensional data positions, which encourages them to encode information about high dimensional data proximity.

Let $P$ be a distribution over directed edges in a kNN graph $P$, and let $ij$ denote an edge from head node $i$ to tail node $j$ in the kNN graph. 
Let $\theta$ be a learned matrix of low dimensional positions.
Let $q$ be the Cauchy kernel:
\begin{equation}
q(\theta_i, \theta_j) =\frac{1}{1 + ||\theta_i - \theta_j||^2}
\end{equation}
where $\theta_i$ denotes the $i$th row of $\theta$.
Following the notation in \citet{damrich2023tsneumapcontrastivelearning}, we will often write $q(ij)$ as a stand in for $q(\theta_i, \theta_j)$ for notational convenience.
Let $\xi$ be a uniform noise distribution over all edges in the complete digraph over nodes in the kNN graph, from which we can draw noise samples, $m \sim \xi$.
Note that we will occasionally write $M \sim \xi$ to represent several samples drawn i.i.d from $\xi$.
The InfoNC-t-SNE loss function is written:
\begin{equation}
\mathcal{L} = -\mathbb{E}_{\substack{ij \sim P \\ M \sim \xi}} \left[ \log \left( \frac{q(ij)}{q(ij) + \sum_{m \in M} q(im)} \right) \right] 
\label{eq:InfoNCELoss}
\end{equation}

As outlined in \citet{damrich2023tsneumapcontrastivelearning}, the minima of $\mathcal L$, which we denote $\tilde \theta$, have the beneficial property that $q_{\tilde \theta} = P$, so long as $\xi$ has full support and there exists some $\theta^*$ where $q_{\theta ^*}=P$.
Intuitively, this indicates that optimizing $\mathcal L$ will recover $P$, so long as the class of functions being optimized over is expressive enough to represent $P$, and the selected noise distribution has full support.

\subsection{NOMAD Projection: Positive Force Calculation}
Most modern data mapping algorithms rely on an external approximate nearest neighbor (ANN) algorithm when constructing a kNN graph from vector data.
Popular choices for ANN construction, including FAISS \citep{douze2024faisslibrary} and PyNNDescent \citep{mcinnes2020umapuniformmanifoldapproximation}, do not cleanly shard across multiple devices, which impedes distributed training.
In particular, cases where neighbors in the kNN graph reside on different devices force inter-device communication during positive spring force calculation.

To remedy this, NOMAD Projection opts for a simple K-Means based ANN index.
We initialize our K-Means clustering using a locally sensitive hash, run expectation maximization until convergence, and compute exact nearest neighbors for each point within its cluster.
Since the only candidates considered for a target point's neighbors share a cluster with the target point, each cluster is a component of the resulting ANN graph.
This enables us to shard clusters across GPUs while ensuring that every point's approximate nearest neighbors remain on the same device, circumventing the need for inter-device communication during positive spring force calculation.
This distribution strategy is shown in Figure \ref{fig:distribution}.

\subsection{NOMAD Projection: Negative Force Calculation}
\label{sec:negforce}

\begin{figure*}[ht]
    \centering
    \begin{subfigure}{0.49\linewidth}
        \includegraphics[width=1.0\linewidth]{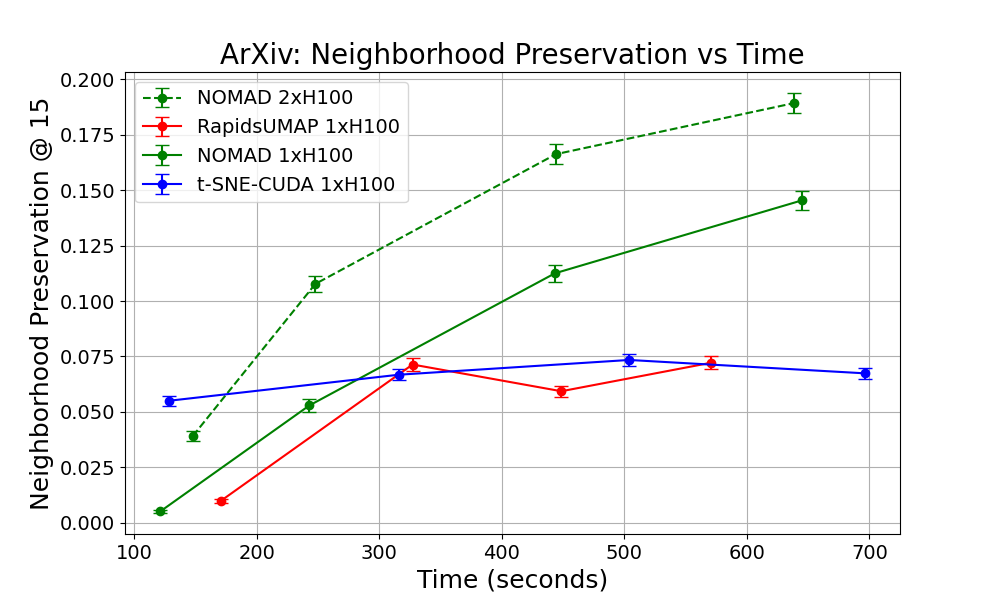}
    \end{subfigure}
    \hfill
    \begin{subfigure}{0.49\linewidth}
        \includegraphics[width=1.0\linewidth]{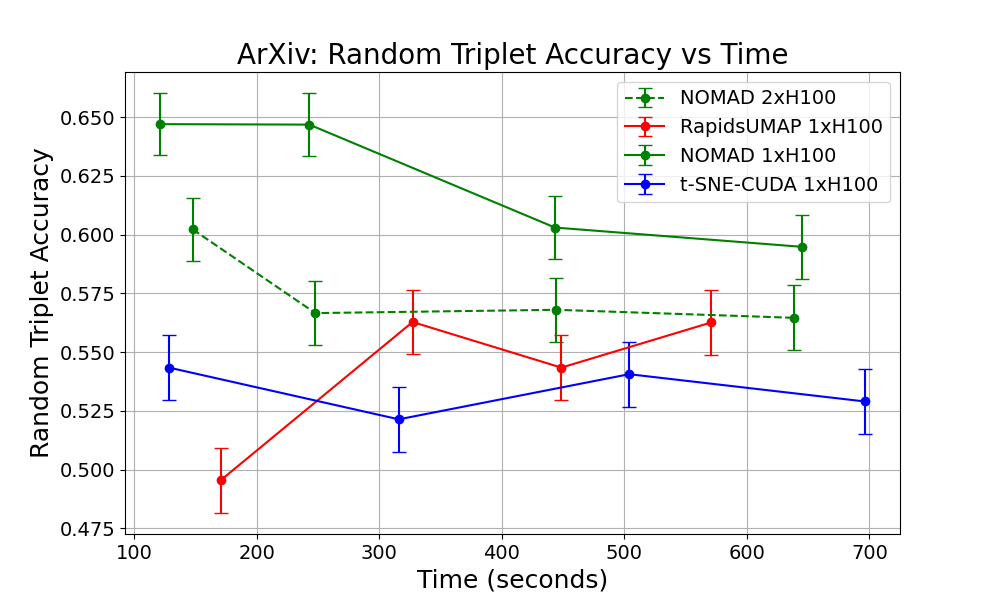}
    \end{subfigure}
    \vskip\baselineskip
    \begin{subfigure}{0.49\linewidth}
        \includegraphics[width=1.0\linewidth]{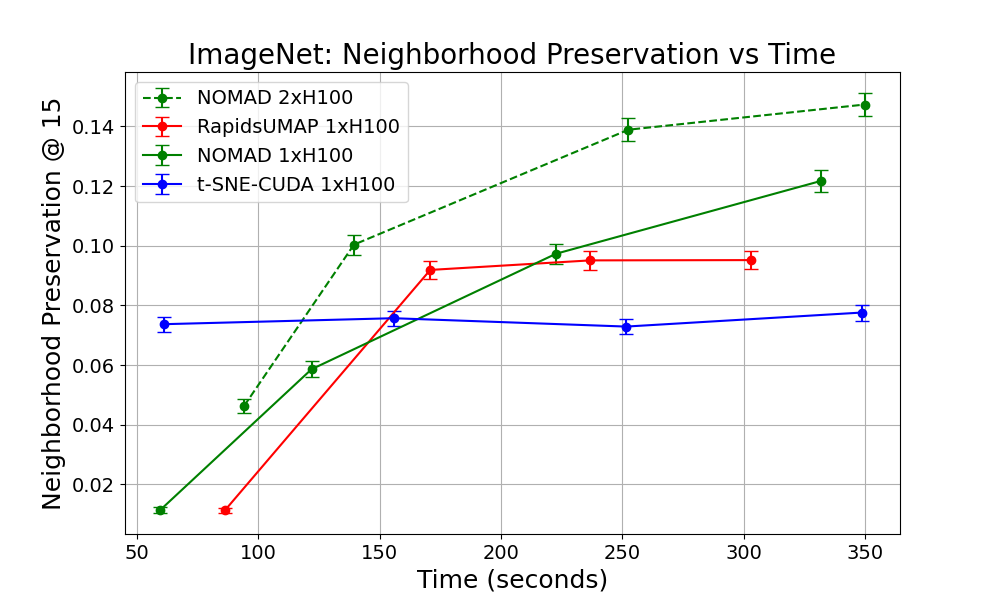}
    \end{subfigure}
    \hfill
    \begin{subfigure}{0.49\linewidth}
        \includegraphics[width=1.0\linewidth]{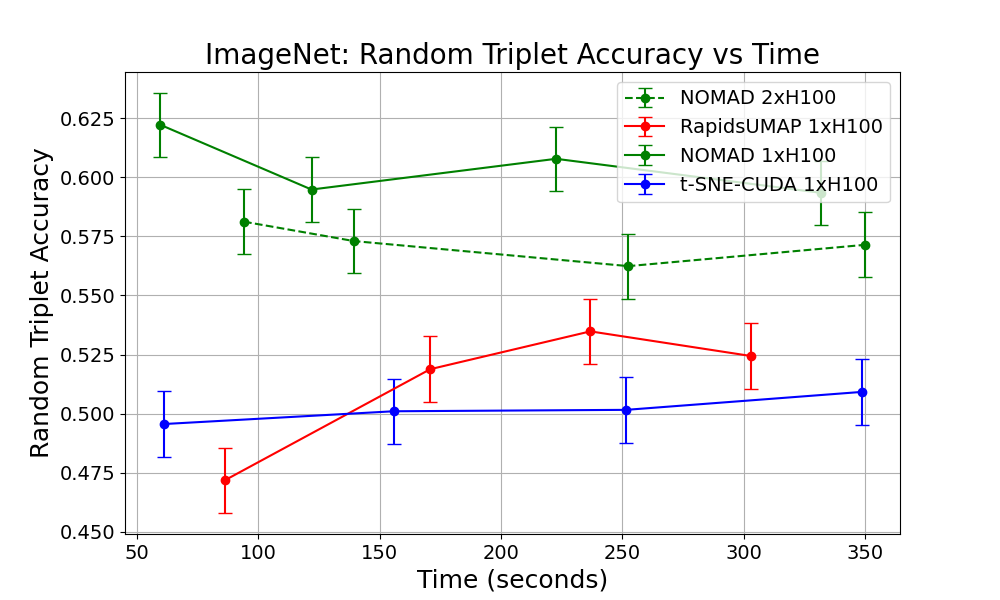}
    \end{subfigure}
    \caption{A comparison of the speed and performance of several GPU accelerated data mapping algorithms. In all cases, NOMAD Projection achieves similar or superior neighborhood preservation and random triplet accuracy to existing methods when run for sufficiently many epochs. It is worth noting that t-SNE-CUDA achieves impressively fast neighborhood preservation scores, but struggles to continue improving with additional epochs of training. NOMAD Projection can take advantage of multiple GPUs to significantly improve its neighborhood preservation performance at the expense of random triplet accuracy. Note that t-SNE-CUDA does not take advantage of techniques for improving global coherence such as early exaggeration or spectral initialization, which we conjecture negatively impacts its random triplet accuracies.}
    \label{fig:arxiv-imagenet}
\end{figure*}

NOMAD Projection minimizes a surrogate loss \cite{Murphy2012} that approximates an upper bound to the InfoNC-t-SNE loss function. This surrogate loss is constructed to enable efficient communication between multiple devices during negative spring force calculation.
In particular, the NOMAD Projection loss only requires the cluster means to be all-gathered during negative force calculation, drastically reducing the inter-device communication required to compute negative forces. 

Intuitively, the NOMAD Projection loss simply treats cluster means like negative samples during InfoNCE classification.

Consider a partition $R$ of the support of the noise distribution $\xi$.
$R$ will naturally induce a partition over any sample $M \sim \xi$, according to which cell of $R$ each $m \in M$ originates from.
We write $M_r$ to signify the subset of a sample $M$ that originates from cell $r$.
We define $\tilde R$ as the subset of cells in $R$ that we wish to approximate.
The NOMAD Projection loss function is written:
\begin{align}
\mathcal L &= -\mathbb{E}_{\substack{i \sim P_i}} \left[ \sum_{j} p(j|i )\log \left( \frac{q(ij)}{q(ij) + \tilde {\mathcal M} + \mathcal M}\right)\right]
\label{eq:NOMADProjectLossText}
\end{align}
\begin{align}
\tilde {\mathcal M} &= |M| \sum_{r \in \tilde R} p(m \in r) q (i \mu_r )\\
\mathcal M &= \sum_{r \in R\setminus \tilde R} \mathbb{E}_{\substack{M \sim \xi}} \left[ \sum_{m \in M_r} q(im)\right]
\end{align}
where $\mu_r$ is the expected value of the vector corresponding to tails drawn from the conditional noise distribution of cell $r$, denoted $\xi_r$, and $P_i$ is the marginal distribution over heads of directed edges induced by $P$.
Note the slight abuse of notation here, where $q(i\mu_r)$ is taken to denote $q(\theta_i, \mu_r)$.
In the case where we choose not to approximate any partition cells, $\tilde M = 0$, and Equation \ref{eq:NOMADProjectLossText} reduces to the standard InfoNC-t-SNE loss in Equation \ref{eq:InfoNCELoss}.

InfoNC-t-SNE models $p(j|i)$ implicitly when sampling uniformly from the joint distribution $P$.
In contrast, t-SNE models $p(j|i)$ explicitly via its bandwidth hyperparameter.
Like t-SNE, NOMAD Projection explicitly models $p(j|i)$.
Unlike t-SNE, NOMAD Projection uses a simple inverse rank model:
\begin{equation}
p(j|i) = 
\begin{cases}
\frac{e^{1/\text{rank}_j(i)}}{\sum_{j=0}^k e^{1/(j+1)}} & \text{rank}_j(i) < k\\
0 & \text{else}\\
\end{cases}
\end{equation}
where $\text{rank}_j(i)$ returns the index of $i$ in a list of points sorted in ascending order of distance to $j$.
This model is simpler to compute than the data-dependent $p(j|i)$ presented in t-SNE.

NOMAD Projection uses a uniform marginal distribution on edge heads $P_i$, and approximates $\mathbb E_{i \sim P_i}$ via random sampling.
NOMAD Projection also uses a uniform noise distribution over edge tails $\xi$.

\begin{theorem}
Let $G$ be an ANN graph formed from a vector dataset.
Let $P$ be a probability distribution over directed edges in $G$ with finite moments, $\xi$ be a uniform distribution over all edges in the complete digraph on nodes in $G$, $R$ be a partition of the support of $\xi$, and $q$ be the Cauchy kernel.
For all $P$, $G$, and $R$, Equation \ref{eq:NOMADProjectLossText} $\gtrsim$ Equation \ref{eq:InfoNCELoss}.
\label{thm:bound}
\end{theorem}

Please refer to Section \ref{sec:NOMADProjectLossFull} for a full proof of Theorem \ref{thm:bound}.

\subsection{Additional Implementation Details}

We optimize our loss using stochastic gradient descent with a linear learning rate decay \citep{damrich2023tsneumapcontrastivelearning}.
We set our initial learning rate to $n/10$, a factor of $10$ less than the t-SNE convention presented in \citet{belkina2019automated}, as we find it helps with convergence.
In all cases, we linearly anneal this learning rate to $0$ over the course of training.
We initialize our projection with PCA, as it has been found to improve global structure \citep{wang2021understandingdimensionreductiontools}.

\section{Experimental Results}

We compare NOMAD Projection to existing methods by computing visualizations of several corpora of interest to the machine learning community, and measuring their neighborhood preservation and random triplet accuracy.

Neighborhood preservation at $k$ measures the average overlap between $k$-neighborhoods in two spaces.
Measuring the neighborhood preservation between high and low dimensional spaces is a common measure of local structure preservation in dimensionality reduction \citep{damrich2023tsneumapcontrastivelearning,gonzalez2024landscape}.

Random triplet accuracy measures the probability that a random triplet of data will have the same pairwise distance ordering in both the high and low dimensional space.
Following \citet{wang2021understandingdimensionreductiontools}, we use it as a measure of global structure preservation.

\begin{table*}[ht]
\centering
\begin{tabular}{|l|c|c|c|c|}
\hline
\textbf{Method} & \textbf{Compute} & \textbf{NP@10} & \textbf{Time (hours)} & \textbf{Speedup} \\ \hline
OpenTSNE \cite{Policar2024}       & 16 x 2.9GHz Xeon CPUs & 6.2\%           & 8.0                  & 1x               \\ \hline
NOMAD Projection        &8 x H100 GPUs&6.1\% $\pm$ 0.3\%          & 1.47                  & 5.4x             \\ \hline
RapidsUMAP \cite{nolet2021bringingumapcloserspeed}        &  1 x H100 GPU &-        & OOM                 & -             \\ \hline
t-SNE-CUDA \cite{chan2018tsnecudagpuacceleratedtsneapplications}        & 1 x H100 GPU &-           & OOM                  & -             \\ \hline
\end{tabular}
\vspace{0.5cm}
\caption{Speed and performance metrics for several data mapping methods on the PubMed corpus. NOMAD Projection achieves comparable Neighborhood Preservation at 10 (NP@10) to OpenTSNE, and executes significantly faster, since OpenTSNE cannot leverage GPU compute. GPU based implementations RapidsUMAP and t-SNE-CUDA are unable to complete the task due to out of memory (OOM) errors and the inability to scale to multiple GPUs. OpenTSNE metrics are taken from \cite{gonzalez2024landscape}, which does not report standard error.}
\label{tab:pubmed}
\end{table*}

\subsection{ArXiv and ImageNet}

We compare the wall-time, neighborhood preservation, and random triplet accuracy of NOMAD Projection to state-of-the-art GPU accelerated implementations of t-SNE and UMAP; namely, t-SNE-cuda \citep{chan2018tsnecudagpuacceleratedtsneapplications} and NVIDIA RapidsUMAP \citep{nolet2021bringingumapcloserspeed}.
We select ArXiv and ImageNet as our benchmark datasets due to their size and ubiquity in the literature.

We obtain high dimensional vectors for ArXiv using Nomic Embed \citep{nussbaum2024nomicembedtrainingreproducible} and high dimensional vectors for ImageNet using OpenClip \citep{ilharco_gabriel_2021_5143773}.
Our results are presented in Figure \ref{fig:arxiv-imagenet}.
On both datasets, we find that NOMAD Projection achieves similar or superior neighborhood preservation and random triplet accuracy to existing methods when run for sufficiently many epochs.

We also find that NOMAD Projection can take advantage of multiple GPUs to significantly improve its speed and neighborhood preservation performance at the expense of a slight decline in random triplet accuracy.
We believe this trade-off between local and global structure preservation is due to the nature of our partition approximation, as partitioning the ANN graph into multiple components removes positive spring forces that encode information about how each component fits together on a more global scale.
Critically, multi-GPU NOMAD Projection still achieves comparable or superior random triplet accuracy to existing GPU accelerated methods.

\subsection{PubMed}
Previously, the largest published data map was a visualization of the entirety of the PubMed corpus  \citep{gonzalez2024landscape}.
In this map, PubMed documents are vectorized with a custom BERT model, and then reduced to 2 dimensions using OpenTSNE \citep{Policar2024}, a multi-core CPU implementation of FIt-SNE \citep{Linderman_2019}.
OpenTSNE was chosen due to its ability to handle the massive PubMed dataset, a task that required a machine with 0.5 TB of RAM.
OpenTSNE achieved a Neighborhood Preservation at 10 score of $6.2\%$ after 8 hours of computation on the PubMed vectors.
On the same vectors, NOMAD Projection achieves a comparable Neighborhood Preservation score of $6.1 \pm 0.3\%$ in under 1.5 hours of computation.
Other GPU accelerated data mapping methods are unable to complete the task due to out of memory errors on one GPU and the inability to scale to multiple GPUs.
These results are presented in Table \ref{tab:pubmed}.

\subsection{Multilingual Wikipedia}
\label{sec:wikitext}

\begin{figure*}
    \centering
    \includegraphics[]{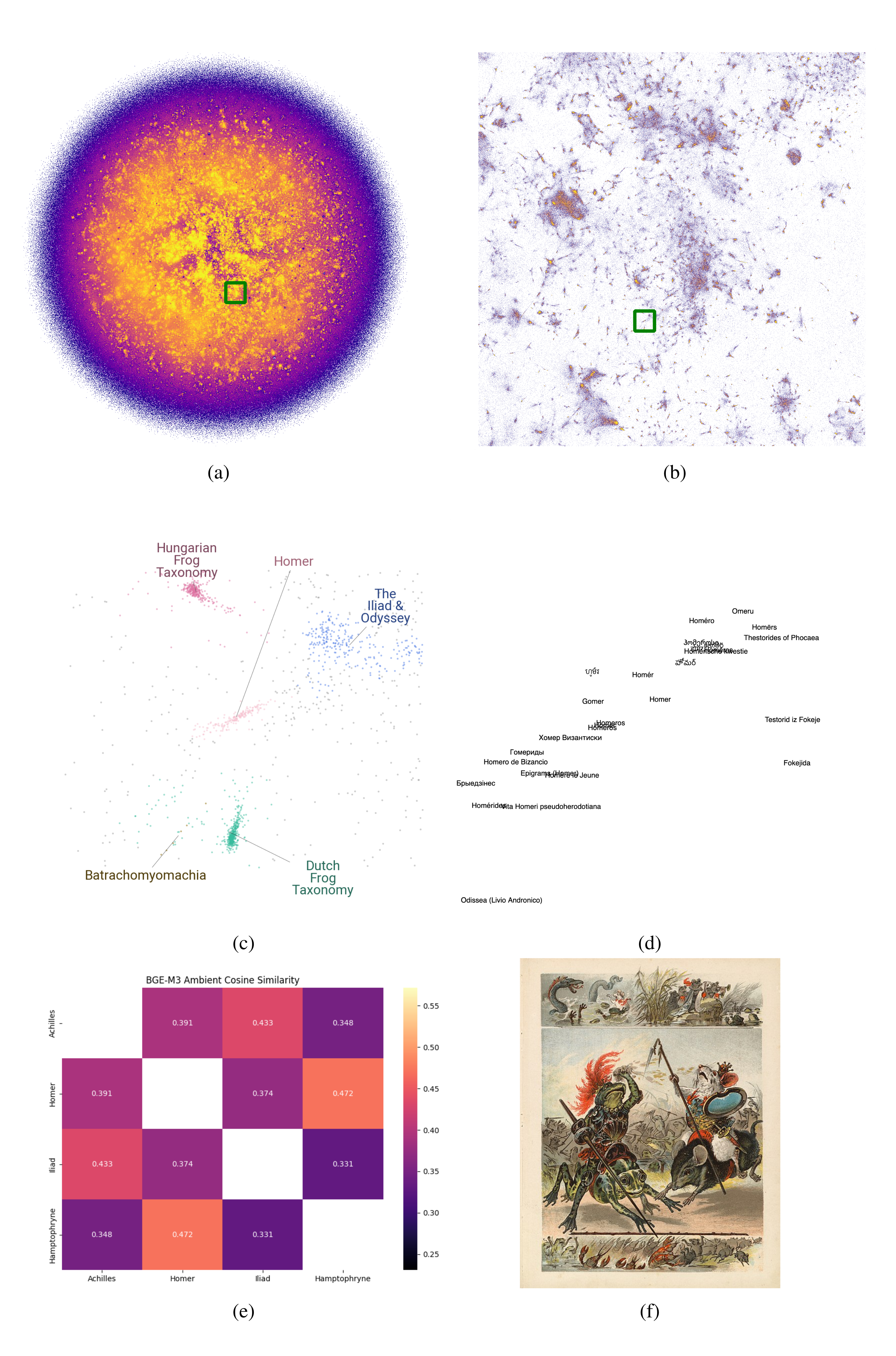}
    \caption{A multiscale qualitative exploration of a NOMAD Projection of Multilingual Wikipedia. A detailed analysis is presented in Section \ref{sec:wikitext}}
    \label{fig:wiki}
\end{figure*}

To fully showcase the scalability of NOMAD Projection, we compute a visualization of the entirety of Multilingual Wikipedia, as shown in Figure \ref{fig:teaser}.
We vectorize Multilingual Wikipedia using the BGE-M3 
 embedding model \citep{chen2024bgem3embeddingmultilingualmultifunctionality},
compute the visualization on an NVIDIA 8xH100 node, and run the optimization for $\sim5.8$ hours.

Figure \ref{fig:wiki} presents a qualitative investigation of the Multilingual Wikipedia map.
In particular, it showcases NOMAD Projection's ability to capture structure at several different levels of zoom. 

Figure \ref{fig:wiki}(a) shows a global view of the Multilingual Wikipedia map. The green rectangle in \ref{fig:wiki}(a) specifies a region populated with articles regarding Greek Mythology. 

Figure \ref{fig:wiki}(b) shows a 20x magnification of the region specified in \ref{fig:wiki}(a). Local clustering structure remains clearly visible at this zoom level. The green rectangle in \ref{fig:wiki}(b) specifies a region populated with articles relating to Homer. 

Figure \ref{fig:wiki}(c) shows a 20$\times$ magnification of the region specified in \ref{fig:wiki}(b).
Figure \ref{fig:wiki}(c) is annotated with titles for 4 clusters present in this portion of the map. 
Two of the clusters (Homer and The Iliad \& Odyssey) relate straightforwardly to Greek Mythology, while 2 clusters (Hungarian Frog Taxonomy and Dutch Frog Taxonomy) are seemingly unrelated.

An investigation of the ambient BGE-M3 vectors, shown in Figure \ref{fig:wiki}(e), reveals that the embedding model itself contains a counterintuitive relationship between frog taxonomy and Greek Mythology.
In particular, \ref{fig:wiki}(e) illustrates the counterintuitive fact that BGE-M3 considers Homer to be more related to Hamptophyrne (a frog in the Dutch Frog Taxonomy cluster) than to the Iliad, one of his most famous works.
We conjecture this relationship exists in the BGE-M3 vectors due to a category error that finds the Greek-derived morphology of taxonomic names to be similar to the Greek content designated by articles like Homer and Iliad.
The presence of this relationship in the ambient BGE-M3 vectors showcases NOMAD projection's ability to effectively surface counterintuitive embedding relationships in its visualizations.

Also specified in \ref{fig:wiki}(c) is the location of ``Batrachomyomachia" articles in several languages.
``Batrachomyomachia" translates to ``Battle of Frogs and Mice," and is a widely translated parody of the Iliad featuring frogs and mice as the main characters.
Figure \ref{fig:wiki}(f) shows an illustration from a Batrachomyomachia Wikipedia page, where frogs and mice clash in the book's analogue of the Battle for Troy.
Batrachomyomachia's placement between Homer and Dutch Frog Taxonomy showcases NOMAD projection's ability to situate content with multiple themes effectively.

Figure \ref{fig:wiki}(d) shows a 5$\times$ magnification of the Homer cluster in \ref{fig:wiki}(c).
A random sample of 25 article titles have been plotted at their corresponding map locations. 
These articles overwhelmingly make reference to Homer in various languages, showcasing NOMAD Projection's ability to maintain coherence on extremely local scales.

Overall, the investigation presented in Figure \ref{fig:wiki} demonstrates that NOMAD Projection effectively captures relationships across several scales in its layouts.

\section{Conclusion}

In this work, we introduced NOMAD Projection, an unstructured data visualization method that enables distributed computation by approximating an upper bound on the InfoNC-t-SNE loss.
We provide empirical results that demonstrate NOMAD Projection's superior performance and speed profile when compared to existing data mapping methods on the ArXiv, ImageNet, and PubMed datasets.
We also demonstrate that NOMAD Projection can effectively take advantage of multiple GPUs at train time to alleviate the vRAM cap of existing GPU accelerated data mapping methods.
This enables us to compute the largest data map ever made, a 60 million point map of Multilingual Wikipedia.

\section{Limitations \& Future Work}

There are several clear next steps for this line of research.
We have only explored NOMAD Projection as a distribution strategy for several GPUs on a single node.
The obvious next step is to adapt NOMAD Projection for use on multi-node networks. 
A multi-node implementation of NOMAD Projection would require a nontrivial extension of our method, particularly when accounting for the differences in bandwidth and latency profiles between inter-node and intra-node GPU communication.

Additionally, the NOMAD Projection code remains largely unoptimized.
There are likely several significant speedups that can be unlocked through code profiling.

Further, we believe that NOMAD Projection's theoretical relationship to InfoNCE can be applied in areas extending beyond data mapping, including more general contrastive learning and language modeling settings.

Finally, we acknowledge that there is additional characterization of NOMAD Projection to be done beyond the experiments provided in this paper.
We provide open source code and datasets with which our methods can be replicated to facilitate this work.

\newpage

\section{NOMAD Projection Approximates an Upper Bound on InfoNC-t-SNE}
\label{sec:NOMADProjectLossFull}
For this proof, we adopt the notational conventions and assumptions outlined in Section \ref{sec:negforce}.
Let $\mathcal {L}^{I}$ be the InfoNC-t-SNE loss.
We start by expressing the sum over negative samples in the vanilla InfoNC-t-SNE in terms of our partition:
\setcounter{equation}{0}
\begin{align}
\centering
 \mathcal{L}^{I}&= -\mathbb{E}_{\substack{ij \sim P \\ M \sim \xi}} \left[ \log \left( \frac{q(ij)}{q(ij) + \sum_{m \in M} q(im)} \right) \right] \\
&= -\mathbb{E}_{\substack{ij \sim P \\ M \sim \xi}} \left[ \log \left( \frac{q(ij)}{q(ij) + \sum\limits_{r \in R} \sum\limits_{m \in M_r} q(im)  } \right) \right]
\end{align}
We then isolate the terms corresponding to negative samples and focus on approximately bounding them:
\begin{align}
&= -\mathbb{E}_{\substack{ij \sim P \\ M \sim \xi}} \bigg[ \log \left( q(ij) \right) \bigg] \nonumber \\ &+ \mathbb{E}_{\substack{ij \sim P \\ M \sim \xi}} \left[\log \left( q(ij) + \sum_{r \in R }\sum_{m \in M_r} q(im) \right) \right]
\end{align}
Applying linearity of expectation:
\begin{align}
&\mathbb{E}_{\substack{ij \sim P \\ M \sim \xi}} \left[\log \left( q(ij) + \sum_{r \in R }\sum_{m \in M_r} q(im) \right) \right]\\
&=  \mathbb{E}_{\substack{ij \sim P}} \left[\mathbb{E}_{\substack{M \sim \xi}} \left[ \log \left( q(ij) + \sum_{r \in R }\sum_{m \in M_r} q(im) \right) \right] \right]
\end{align}
Since $\log$ is concave, we can apply Jensen's inequality:
\begin{align}
\label{pf:jensen1}
&\leq \mathbb{E}_{\substack{ij \sim P}} \left[ \log \left( \mathbb{E}_{M \sim \xi} \left[q(ij) + \sum_{r \in R }\sum_{m \in M_r} q(im) \right] \right)\right]\\
&= \mathbb{E}_{\substack{ij \sim P}} \left[ \log \left( q(ij) + \sum_{r \in R } \mathbb{E}_{M \sim \xi} \bigg [ \sum_{m \in M_r} q(im) \bigg ] \right) \right]\\
&= \mathbb{E}_{\substack{ij \sim P}} \left[ \log \left( q(ij) + \sum_{r \in R } \mathbb{E}_{M \sim \xi} \bigg [ \sum_{m \in M} q(im) \mathbb I_{m \in r} \bigg ] \right) \right]
\end{align}
where $\mathbb I_{m \in r}$ is the indicator that a sample $m$ originates from cell $r$.
Since all the $m$ are i.i.d.:
\begin{align}
&= \mathbb{E}_{\substack{ij \sim P}} \left[ \log \left( q(ij) + \sum_{r \in R } |M| \mathbb{E}_{m \sim \xi} \bigg [ q(im) \mathbb I_{m \in r} \bigg ] \right) \right]\\
&= \mathbb{E}_{\substack{ij \sim P}} \left[ \log \left( q(ij) + |M| \sum_{r \in R } p(m \in r) \mathbb{E}_{m \sim \xi_r} \bigg [ q(im)  \bigg ] \right) \right]
\label{eq:presub}
\end{align}
where $\xi_r$ denotes the distribution of $\xi$ conditioned on a sample originating from cell $r$.
Recall that $q(im)$ is shorthand for $q(\theta_i, \theta_m)$.
Applying a first order Taylor Expansion to $q(\theta_i, \theta_m)$ about $\theta_m = \mathbb E_{m \sim \xi_r}[\theta_m] = \mu_r$, we have:
\begin{align}
\mathbb E_{m \sim \xi_m} [q(im)] \approx q(i\mu_r)
\end{align}
where $q(i\mu_r)$ denotes $q(\theta_i, \mu_r)$.
Note that the linear terms in the Taylor Expansion vanish in expectation, so the approximation is accurate up to the second order.
Thus:
\begin{align}
\mathbb{E}_{\substack{ij \sim P}} \left[ \log \left( q(ij) + |M| \sum_{r \in R } p(m \in r) \mathbb{E}_{m \sim \xi_r} \bigg [ q(im)  \bigg ] \right) \right]\\
\approx \mathbb{E}_{\substack{ij \sim P}} \left[ \log \left( q(ij) + |M| \sum_{r \in R } p(m \in r)  q(i\mu_r)\right) \right]
\end{align}

The key result of this derivation is that replacing noise samples with the appropriately weighted mean of the partition they belong to is equivalent to approximating an upper bound on the InfoNC-t-SNE loss.

Without loss of generality, this approximation can be applied to any subset of the cells in the partition, yielding:

\begin{align}
\mathcal{L}^{I} & \lesssim -\mathbb{E}_{\substack{ij \sim P}} \left[\log \left( \frac{q(ij)}{q(ij) + \tilde {\mathcal M} + \mathcal M}\right)\right]
\label{eq:NOMADProjectLoss}
\end{align}
where
\begin{align}
\tilde {\mathcal M} &= |M| \sum_{r \in \tilde R} p(m \in r) q (i \mu_r )\\
\mathcal M &= \sum_{r \in R\setminus \tilde R} \mathbb{E}_{\substack{M \sim \xi}} \left[ \sum_{m \in M_r} q(im)\right]
\end{align}

We apply the definition of expectation to yield the final expression:
\begin{align}
\mathcal{L}^{I} &\lesssim -\mathbb{E}_{\substack{i \sim P_i}} \left[ \sum_{j} p(j|i) \log \left( \frac{q(ij)}{q(ij) + \tilde {\mathcal M} + \mathcal M}\right)\right]
\label{eq:NOMADProjectLoss}
\end{align}

\newpage

{
    \small
    \bibliographystyle{ieeenat_fullname}
    \bibliography{main}
}


\end{document}